\documentclass[conference]{IEEEtran}
\usepackage[utf8]{inputenc}
\usepackage{amsmath,amssymb,amsfonts,bm}
\usepackage{algorithmic}
\usepackage{graphicx}
\usepackage{textcomp}
\usepackage{xcolor}
\usepackage{tikz}
\usepackage{hyperref}
\usepackage{cite}
\usepackage{booktabs}

\begin{document}

\title{\huge Next Generation Multitarget Trackers: Random Finite Set Methods vs Transformer-based Deep Learning}
\author{Juliano Pinto, Georg Hess, William Ljungbergh, Yuxuan Xia, Lennart Svensson, Henk Wymeersch\\
Department of Electrical Engineering, Chalmers University of Technology, Sweden
\\
Email: juliano@chalmers.se}

\maketitle

\begin{abstract}
Multitarget Tracking (MTT) is the problem of tracking the states of an unknown number of objects using noisy measurements, with important applications to autonomous driving, surveillance, robotics, and others.
In the model-based Bayesian setting, there are conjugate priors that enable us to express the multi-object posterior in closed form, which could theoretically provide Bayes-optimal estimates. 
However, the posterior involves a super-exponential growth of the number of hypotheses over time, forcing state-of-the-art methods to resort to approximations for remaining tractable, which can impact their performance in complex scenarios.
Model-free methods based on deep-learning provide an attractive alternative, as they can, in principle, learn the optimal filter from data, but to the best of our knowledge were never compared to current state-of-the-art Bayesian filters, specially not in contexts where accurate models are available.
In this paper, we propose a high-performing deep-learning method for MTT based on the Transformer architecture and compare it to two state-of-the-art Bayesian filters, in a setting where we assume the correct model is provided. Although this gives an edge to the model-based filters, it also allows us to generate unlimited training data. We show that the proposed model outperforms state-of-the-art Bayesian filters in complex scenarios, while matching their performance in simpler cases, which validates the applicability of deep-learning also in the model-based regime. The code for all our implementations is made available at \url{https://github.com/JulianoLagana/MT3}.
\end{abstract}


\section{Introduction}

Multitarget tracking (MTT) is the problem of tracking a varying number of targets/objects through time based on noisy measurements. Methods capable of such functionality are important for several applications, such as pedestrian tracking \cite{mot_in_pedestrian_tracking1}, autonomous driving \cite{mot_in_autonomous_driving1}, air traffic control \cite{mot_in_air_traffic_control}, oceanography \cite{mot_in_oceanography}, and many others. The main challenge of MTT comes from the fact that objects may appear and disappear from the scene at every time-step, and that there can be missed and false detections. Therefore, a key challenge of MTT is to address the unknown correspondence between targets and measurements, a task referred to as data association. 

In situations where we have access to accurate multitarget models, and observe only low-dimensional, individual object detections, the state of the art is achieved by model-based Bayesian methods. Typical examples are the Poisson multi-Bernoulli mixture (PMBM) filter \cite{PMBM} and the delta-generalized labeled multi-Bernoulli ($\delta$-GLMB) \cite{GLMB} filter, which use the random finite set framework to formulate the tracking problem. These methods make use of multi-object conjugate priors to obtain a closed-form expression for the multi-object posterior and can, in theory, provide Bayes-optimal estimates. In practice, however, the complexity of the data association and track management makes it intractable to compute the optimal solution \cite{PMBM,GLMB}. Thus, these methods must resort to approximations of the true posterior density, which become increasingly inaccurate as the data association problem becomes more challenging.

In recent years, MTT algorithms based on deep learning have emerged as attractive alternatives to traditional Bayesian methods, usually optimizing a model with a large number of parameters by minimizing the empirical risk on an annotated dataset for the problem at hand \cite{deep_learning_book}. Such methods have achieved unmatched performance in settings where no models are available and high-dimensional measurements are provided \cite{mot_challenge, deep_learning_video_tracking_survey}, by extracting informative features from the data to aid in estimating the quantities of interest. However, to the best of our knowledge these methods have never been directly compared to the performance of current state-of-the-art model-based Bayesian methods such as the PMBM and $\delta$-GLMB filters, specially not in contexts where accurate models of the environment are available.

In this paper, we study how deep learning MTT compares to the state-of-the-art Bayesian filters PMBM and $\delta$-GLMB, in the model-based regime. In particular, we propose the MultiTarget Tracking Transformer (MT3), a high-performing, specific type of neural network for MTT based on the Transformer \cite{transformer_paper} architecture, and evaluate it in scenarios where we assume that the models are correct. In principle, this context gives the model-based methods an edge because they have access to the true models. However, at the same time we can use the available models to generate as much training data as necessary for MT3, allowing high-capacity architectures that can potentially obtain better performance. Our results show that MT3 performs competitively to PMBM and $\delta$-GLMB with reasonable computational complexity in a relatively simple scenario, while outperforming them when the task becomes more complex. This demonstrates the applicability of deep-learning-based MTT methods also in the model-based regime.

The remainder of the paper is organized as follows. In Section II, we present the multitarget models used and the problem formulation. In Section III, a short background on the Transformer architecture is provided. Section IV details the proposed MT3 algorithm, and Section V the simulation results and ablation studies, followed by conclusions in Section VI.

\subsubsection*{Notations}
Throughout the paper, we use the following notations: scalars are denoted by lowercase or uppercase letters with no special typesetting $x$, vectors by lowercase boldface letters $\mathbf x$, matrices by uppercase boldface letters $\mathbf X$, and sets by uppercase blackboard-bold letters $\mathbb X$. Sequences of scalars are abbreviated as $x_{1:n}$, and sequences of vectors as $\mathbf x_{1:n}$. In addition, we define $\mathbb N_a = \{i \in \mathbb N~|~i\leq a\}, a\in\mathbb N$. 

\section{Multitarget Models and Problem Formulation}
\label{sec:problem_formulation}

In this work we follow the standard multitarget transition and observation models for point objects \cite[Chap. 5]{mahler2014}, without target spawning. The state vector for object $i$ at time-step $t$ is $\mathbf{x}^t_i \in \mathbb R^{d_x}$, and $\mathbb X^t$ is the set of all object states at time-step $t$. New objects arrive according to a Poisson point process (PPP) with birth intensity $\lambda_b(\mathbf x)$, while existing objects depart according to iid Markovian processes where the survival probability is $p_s(\mathbf x)$. The motion models for the objects are iid Markovian processes, where the single-object transition density is denoted $f(\mathbf x^{t+1} ~|~ \mathbf x^t)$. 

At every time-step, each of the existing objects may give rise to at most one measurement (and each measurement is the result of at most one object), where the probability of detection in state $\mathbf x$ is $p_d(\mathbf x)$. Clutter measurements arrive according to a PPP with intensity $\lambda_c$, independent of the existing objects or true measurements. Each non-clutter/true measurement is independent of all other objects and measurements conditioned on its corresponding target, and the single object measurement likelihood is denoted $\mathbf g(\mathbf z^t~|~\mathbf x^t)$, $\mathbf z^t\in\mathbb R^{d_z}$. Lastly, the set of all measurements (true and clutter) generated at time-step $t$ is denoted $\mathbb{Z}^t$. For this paper we focus on the problem of multitarget estimation using a moving window: the estimation of $\mathbb X^T$ given knowledge of the measurements from $\tau$ time-steps in the past until the current time $[\mathbb Z_{T-\tau}, \cdots, \mathbb Z_T]$.

\section{Background on Transformers}
\begin{figure}
    \centering

\tikzset{every picture/.style={line width=0.75pt}} 

\begin{tikzpicture}[x=0.75pt,y=0.75pt,yscale=-1,xscale=1]

\draw    (105.67,330.71) -- (105.67,290.33) ;
\draw [shift={(105.67,287.33)}, rotate = 450] [fill={rgb, 255:red, 0; green, 0; blue, 0 }  ][line width=0.08]  [draw opacity=0] (7.14,-3.43) -- (0,0) -- (7.14,3.43) -- (4.74,0) -- cycle    ;
\draw    (55,304.76) -- (105.67,304.76) ;
\draw    (55,232.76) -- (55,304.76) ;
\draw    (55,232.76) -- (70.57,232.76) ;
\draw [shift={(73.57,232.76)}, rotate = 180] [fill={rgb, 255:red, 0; green, 0; blue, 0 }  ][line width=0.08]  [draw opacity=0] (7.14,-3.43) -- (0,0) -- (7.14,3.43) -- (4.74,0) -- cycle    ;
\draw    (105.67,222.22) -- (105.67,204.89) ;
\draw [shift={(105.67,201.89)}, rotate = 450] [fill={rgb, 255:red, 0; green, 0; blue, 0 }  ][line width=0.08]  [draw opacity=0] (7.14,-3.43) -- (0,0) -- (7.14,3.43) -- (4.74,0) -- cycle    ;
\draw    (106.67,257.76) -- (106.67,244.9) ;
\draw [shift={(106.67,241.9)}, rotate = 450] [fill={rgb, 255:red, 0; green, 0; blue, 0 }  ][line width=0.08]  [draw opacity=0] (7.14,-3.43) -- (0,0) -- (7.14,3.43) -- (4.74,0) -- cycle    ;
\draw    (55,158.76) -- (69.64,158.76) ;
\draw [shift={(72.64,158.76)}, rotate = 180] [fill={rgb, 255:red, 0; green, 0; blue, 0 }  ][line width=0.08]  [draw opacity=0] (7.14,-3.43) -- (0,0) -- (7.14,3.43) -- (4.74,0) -- cycle    ;
\draw    (55,158.76) -- (55,214.06) ;
\draw    (55,214.06) -- (105.67,214.06) ;
\draw    (106.67,182.9) -- (106.67,170.9) ;
\draw [shift={(106.67,167.9)}, rotate = 450] [fill={rgb, 255:red, 0; green, 0; blue, 0 }  ][line width=0.08]  [draw opacity=0] (7.14,-3.43) -- (0,0) -- (7.14,3.43) -- (4.74,0) -- cycle    ;
\draw  [dash pattern={on 4.5pt off 4.5pt}] (41.89,162.49) .. controls (41.89,150.19) and (51.86,140.22) .. (64.16,140.22) -- (130.96,140.22) .. controls (143.25,140.22) and (153.22,150.19) .. (153.22,162.49) -- (153.22,295.45) .. controls (153.22,307.75) and (143.25,317.71) .. (130.96,317.71) -- (64.16,317.71) .. controls (51.86,317.71) and (41.89,307.75) .. (41.89,295.45) -- cycle ;
\draw  [fill={rgb, 255:red, 252; green, 224; blue, 180 }  ,fill opacity=1 ] (72.22,263.51) .. controls (72.22,260.22) and (74.89,257.56) .. (78.18,257.56) -- (135.21,257.56) .. controls (138.5,257.56) and (141.17,260.22) .. (141.17,263.51) -- (141.17,281.38) .. controls (141.17,284.67) and (138.5,287.33) .. (135.21,287.33) -- (78.18,287.33) .. controls (74.89,287.33) and (72.22,284.67) .. (72.22,281.38) -- cycle ;
\draw  [fill={rgb, 255:red, 255; green, 113; blue, 113 }  ,fill opacity=0.3 ] (73.67,226.08) .. controls (73.67,223.95) and (75.39,222.22) .. (77.53,222.22) -- (137.36,222.22) .. controls (139.49,222.22) and (141.22,223.95) .. (141.22,226.08) -- (141.22,237.66) .. controls (141.22,239.79) and (139.49,241.52) .. (137.36,241.52) -- (77.53,241.52) .. controls (75.39,241.52) and (73.67,239.79) .. (73.67,237.66) -- cycle ;
\draw  [fill={rgb, 255:red, 177; green, 245; blue, 249 }  ,fill opacity=1 ] (88.22,187.09) .. controls (88.22,184.95) and (89.95,183.22) .. (92.09,183.22) -- (121.3,183.22) .. controls (123.44,183.22) and (125.17,184.95) .. (125.17,187.09) -- (125.17,198.69) .. controls (125.17,200.82) and (123.44,202.56) .. (121.3,202.56) -- (92.09,202.56) .. controls (89.95,202.56) and (88.22,200.82) .. (88.22,198.69) -- cycle ;
\draw    (237.33,333.71) -- (237.33,288.9) ;
\draw [shift={(237.33,285.9)}, rotate = 450] [fill={rgb, 255:red, 0; green, 0; blue, 0 }  ][line width=0.08]  [draw opacity=0] (7.14,-3.43) -- (0,0) -- (7.14,3.43) -- (4.74,0) -- cycle    ;
\draw    (286,304.33) -- (237.67,304.33) ;
\draw    (286,230.56) -- (286,304.33) ;
\draw    (286,230.56) -- (274.22,230.56) ;
\draw [shift={(271.22,230.56)}, rotate = 360] [fill={rgb, 255:red, 0; green, 0; blue, 0 }  ][line width=0.08]  [draw opacity=0] (7.14,-3.43) -- (0,0) -- (7.14,3.43) -- (4.74,0) -- cycle    ;
\draw    (248.67,221.48) -- (248.67,202.14) ;
\draw [shift={(248.67,199.14)}, rotate = 450] [fill={rgb, 255:red, 0; green, 0; blue, 0 }  ][line width=0.08]  [draw opacity=0] (7.14,-3.43) -- (0,0) -- (7.14,3.43) -- (4.74,0) -- cycle    ;
\draw    (237.33,256) -- (237.33,243.14) ;
\draw [shift={(237.33,240.14)}, rotate = 450] [fill={rgb, 255:red, 0; green, 0; blue, 0 }  ][line width=0.08]  [draw opacity=0] (7.14,-3.43) -- (0,0) -- (7.14,3.43) -- (4.74,0) -- cycle    ;
\draw  [dash pattern={on 4.5pt off 4.5pt}] (191.56,68.44) .. controls (191.56,57.09) and (200.76,47.89) .. (212.11,47.89) -- (273.78,47.89) .. controls (285.13,47.89) and (294.33,57.09) .. (294.33,68.44) -- (294.33,297.16) .. controls (294.33,308.51) and (285.13,317.71) .. (273.78,317.71) -- (212.11,317.71) .. controls (200.76,317.71) and (191.56,308.51) .. (191.56,297.16) -- cycle ;
\draw    (106.33,148.57) -- (106.33,141.22) -- (106.33,130.07) ;
\draw    (106.33,130.07) -- (149.22,130.07) -- (171.56,130.07) ;
\draw    (171.56,130.07) -- (171.56,210.64) ;
\draw    (172,210.64) -- (227.22,210.64) ;
\draw    (227.22,210.64) -- (227.22,202.14) ;
\draw [shift={(227.22,199.14)}, rotate = 450] [fill={rgb, 255:red, 0; green, 0; blue, 0 }  ][line width=0.08]  [draw opacity=0] (7.14,-3.43) -- (0,0) -- (7.14,3.43) -- (4.74,0) -- cycle    ;
\draw  [fill={rgb, 255:red, 255; green, 113; blue, 113 }  ,fill opacity=0.3 ] (72.67,152.41) .. controls (72.67,150.28) and (74.39,148.56) .. (76.53,148.56) -- (136.36,148.56) .. controls (138.49,148.56) and (140.22,150.28) .. (140.22,152.41) -- (140.22,163.99) .. controls (140.22,166.12) and (138.49,167.85) .. (136.36,167.85) -- (76.53,167.85) .. controls (74.39,167.85) and (72.67,166.12) .. (72.67,163.99) -- cycle ;
\draw  [fill={rgb, 255:red, 255; green, 113; blue, 113 }  ,fill opacity=0.3 ] (202.67,224.41) .. controls (202.67,222.28) and (204.39,220.56) .. (206.53,220.56) -- (266.36,220.56) .. controls (268.49,220.56) and (270.22,222.28) .. (270.22,224.41) -- (270.22,235.99) .. controls (270.22,238.12) and (268.49,239.85) .. (266.36,239.85) -- (206.53,239.85) .. controls (204.39,239.85) and (202.67,238.12) .. (202.67,235.99) -- cycle ;
\draw  [fill={rgb, 255:red, 252; green, 224; blue, 180 }  ,fill opacity=1 ] (202.89,262.84) .. controls (202.89,259.56) and (205.56,256.89) .. (208.84,256.89) -- (265.88,256.89) .. controls (269.17,256.89) and (271.83,259.56) .. (271.83,262.84) -- (271.83,280.71) .. controls (271.83,284) and (269.17,286.67) .. (265.88,286.67) -- (208.84,286.67) .. controls (205.56,286.67) and (202.89,284) .. (202.89,280.71) -- cycle ;
\draw  [fill={rgb, 255:red, 252; green, 224; blue, 180 }  ,fill opacity=1 ] (198.56,175.18) .. controls (198.56,171.89) and (201.22,169.22) .. (204.51,169.22) -- (273.04,169.22) .. controls (276.33,169.22) and (279,171.89) .. (279,175.18) -- (279,193.04) .. controls (279,196.33) and (276.33,199) .. (273.04,199) -- (204.51,199) .. controls (201.22,199) and (198.56,196.33) .. (198.56,193.04) -- cycle ;
\draw    (286.56,143.67) -- (286.56,159.56) -- (286.56,210.98) ;
\draw    (286.56,143.67) -- (274.57,143.67) ;
\draw [shift={(271.57,143.67)}, rotate = 360] [fill={rgb, 255:red, 0; green, 0; blue, 0 }  ][line width=0.08]  [draw opacity=0] (7.14,-3.43) -- (0,0) -- (7.14,3.43) -- (4.74,0) -- cycle    ;
\draw    (236.67,170) -- (236.67,157.14) ;
\draw [shift={(236.67,154.14)}, rotate = 450] [fill={rgb, 255:red, 0; green, 0; blue, 0 }  ][line width=0.08]  [draw opacity=0] (7.14,-3.43) -- (0,0) -- (7.14,3.43) -- (4.74,0) -- cycle    ;
\draw  [fill={rgb, 255:red, 255; green, 113; blue, 113 }  ,fill opacity=0.3 ] (203.33,138.41) .. controls (203.33,136.28) and (205.06,134.56) .. (207.19,134.56) -- (267.03,134.56) .. controls (269.16,134.56) and (270.89,136.28) .. (270.89,138.41) -- (270.89,149.99) .. controls (270.89,152.12) and (269.16,153.85) .. (267.03,153.85) -- (207.19,153.85) .. controls (205.06,153.85) and (203.33,152.12) .. (203.33,149.99) -- cycle ;
\draw    (248.67,210.98) -- (286.56,210.98) ;
\draw    (235,134.89) -- (235,118.43) ;
\draw [shift={(235,115.43)}, rotate = 450] [fill={rgb, 255:red, 0; green, 0; blue, 0 }  ][line width=0.08]  [draw opacity=0] (7.14,-3.43) -- (0,0) -- (7.14,3.43) -- (4.74,0) -- cycle    ;
\draw    (236,95.57) -- (236,83.57) ;
\draw [shift={(236,80.57)}, rotate = 450] [fill={rgb, 255:red, 0; green, 0; blue, 0 }  ][line width=0.08]  [draw opacity=0] (7.14,-3.43) -- (0,0) -- (7.14,3.43) -- (4.74,0) -- cycle    ;
\draw  [fill={rgb, 255:red, 177; green, 245; blue, 249 }  ,fill opacity=1 ] (217.56,99.76) .. controls (217.56,97.62) and (219.29,95.89) .. (221.42,95.89) -- (250.63,95.89) .. controls (252.77,95.89) and (254.5,97.62) .. (254.5,99.76) -- (254.5,111.36) .. controls (254.5,113.49) and (252.77,115.22) .. (250.63,115.22) -- (221.42,115.22) .. controls (219.29,115.22) and (217.56,113.49) .. (217.56,111.36) -- cycle ;
\draw  [fill={rgb, 255:red, 255; green, 113; blue, 113 }  ,fill opacity=0.3 ] (201,65.08) .. controls (201,62.95) and (202.73,61.22) .. (204.86,61.22) -- (264.7,61.22) .. controls (266.83,61.22) and (268.56,62.95) .. (268.56,65.08) -- (268.56,76.66) .. controls (268.56,78.79) and (266.83,80.52) .. (264.7,80.52) -- (204.86,80.52) .. controls (202.73,80.52) and (201,78.79) .. (201,76.66) -- cycle ;
\draw    (284.22,69) -- (284.22,84.89) -- (284.22,128.56) ;
\draw    (284.22,69) -- (272.07,69) ;
\draw [shift={(269.07,69)}, rotate = 360] [fill={rgb, 255:red, 0; green, 0; blue, 0 }  ][line width=0.08]  [draw opacity=0] (7.14,-3.43) -- (0,0) -- (7.14,3.43) -- (4.74,0) -- cycle    ;
\draw    (235.22,128.56) -- (284.22,128.56) ;
\draw    (236.56,61.58) -- (236.56,35.36) ;
\draw [shift={(236.56,32.36)}, rotate = 450] [fill={rgb, 255:red, 0; green, 0; blue, 0 }  ][line width=0.08]  [draw opacity=0] (7.14,-3.43) -- (0,0) -- (7.14,3.43) -- (4.74,0) -- cycle    ;

\draw (80.17,262) node [anchor=north west][inner sep=0.75pt]  [font=\scriptsize] [align=left] {\begin{minipage}[lt]{38.031448000000005pt}\setlength\topsep{0pt}
\begin{center}
Multi-head \\attention
\end{center}

\end{minipage}};
\draw (79.17,226.76) node [anchor=north west][inner sep=0.75pt]  [font=\scriptsize] [align=left] {Add \& Norm};
\draw (96.09,188.22) node [anchor=north west][inner sep=0.75pt]  [font=\scriptsize] [align=left] {FFN};
\draw (11.33,156.4) node [anchor=north west][inner sep=0.75pt]    {$N\times $};
\draw (293.33,56.3) node [anchor=north west][inner sep=0.75pt]    {$\times M$};
\draw (77.17,153.1) node [anchor=north west][inner sep=0.75pt]  [font=\scriptsize] [align=left] {Add \& Norm};
\draw (208.17,225.1) node [anchor=north west][inner sep=0.75pt]  [font=\scriptsize] [align=left] {Add \& Norm};
\draw (210.83,261) node [anchor=north west][inner sep=0.75pt]  [font=\scriptsize] [align=left] {\begin{minipage}[lt]{38.031448000000005pt}\setlength\topsep{0pt}
\begin{center}
Multi-head \\attention
\end{center}

\end{minipage}};
\draw (204.51,173) node [anchor=north west][inner sep=0.75pt]  [font=\scriptsize] [align=left] {\begin{minipage}[lt]{49.53677600000001pt}\setlength\topsep{0pt}
\begin{center}
Multi-head \\cross-attention
\end{center}

\end{minipage}};
\draw (208.83,139.1) node [anchor=north west][inner sep=0.75pt]  [font=\scriptsize] [align=left] {Add \& Norm};
\draw (225.5,101) node [anchor=north west][inner sep=0.75pt]  [font=\scriptsize] [align=left] {FFN};
\draw (206.86,66.22) node [anchor=north west][inner sep=0.75pt]  [font=\scriptsize] [align=left] {Add \& Norm};
\draw (93.33,334.68) node [anchor=north west][inner sep=0.75pt]    {$\mathbf{x}_{1:n}$};
\draw (225,334.68) node [anchor=north west][inner sep=0.75pt]    {$\mathbf{o}_{1:k}$};
\draw (224.67,19) node [anchor=north west][inner sep=0.75pt]    {$\mathbf{y}_{1:k}$};
\draw (130,117) node [anchor=north west][inner sep=0.75pt]    {$\mathbf{e}_{1:n}$};

\end{tikzpicture}

    \caption{Simplified diagram illustrating the Transformer architecture. Encoder on the left, containing $N$ encoder blocks, processes the input sequence $\mathbf x_{1:n}$ into embeddings $\mathbf e_{1:n}$. DETR decoder on the right, containing $M$ decoder blocks, uses the embeddings $\mathbf e_{1:n}$ produced by the encoder together with the object queries $\mathbf o_{1:k}$ to predict the output sequence $\mathbf y_{1:k}$.}
    \label{fig:encoder-decoder_diagram}
\end{figure}
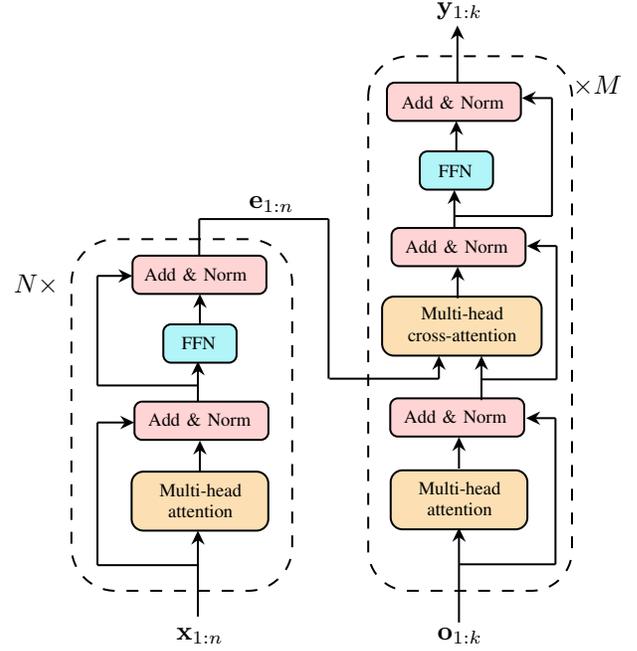

The Transformer architecture, first introduced in \cite{transformer_paper}, is a type of neural network tailored for sequence-to-sequence function approximation, making it well-suited to the MTT problem. Its structure is comprised of an encoder and a decoder, as shown in Fig.\,\ref{fig:encoder-decoder_diagram}. The Transformer processes the input sequence\footnote{The symbols $\mathbf x$, $\mathbf X$, $\mathbf y$, $\mathbf Y$ used for this section are not connected to the rest of the paper; we make use of the standard notation from the Transformer literature (e.g. $\mathbf x$ for the input to layers/modules, and $\mathbf y$ for their outputs).} $\mathbf x_{1:n}$ into an output sequence $\mathbf y_{1:k}$ in a learnable way, and its main building block is the \emph{self-attention layer}. The optimization of the trainable parameters of the Transformer architecture is typically done by stochastic gradient descent on some loss function $\mathcal L_T(\mathbf y_{1:k}, \mathbf g_{1:k})$, that compares the output sequence with a ground-truth sequence $\mathbf g_{1:k}$. 
This section provides a background review on the self-attention layer and the encoder/decoder modules of the Transformer model.

\subsection{Self-Attention Layer}
\label{sec:self-attention}
The self-attention layer is the main building block of the Transformer architecture and processes an input sequence $\mathbf{x}_{1:n}$ into an output sequence $\mathbf y_{1:n}$, where ${\bf x}_i, {\bf y}_i \in \mathbb{R}^d, i \in \mathbb N_n$. We also introduce $\mathbf{X} =\begin{bmatrix}\mathbf x_1, \cdots, \mathbf x_n\end{bmatrix}\in \mathbb R^{d\times n}$ and similarly $\mathbf{Y} = \begin{bmatrix}\mathbf y_1, \cdots, \mathbf y_n\end{bmatrix}\in \mathbb R^{d\times n}$.

The first step is to compute three linear transformations of the input,
\begin{equation}
    \label{eq:self-attention_qkv}
    \mathbf{Q} = \mathbf{W}_Q \mathbf{X},~ \mathbf{K} = \mathbf{W}_K\mathbf{X},~ \mathbf{V}=\mathbf{W}_V\mathbf{X},
\end{equation}
referred to as queries, keys, and values, respectively, and where $\mathbf{W}_Q, \mathbf{W}_K, \mathbf{W}_V \in \mathbb R^{d\times d}$ are the learnable parameters of the self-attention layer. The output is constructed using the queries, keys, and values as
\begin{equation}
    \mathbf{Y}=\mathbf{V}\cdot  \text{Softmax-c}\left(\frac{\mathbf{K}^\top\mathbf{Q}}{\sqrt{d}}\right)~,
\end{equation}
where Softmax-c is the column-wise application of the Softmax function, defined as
\begin{align*}
    &[\text{Softmax-c}(\mathbf Z)]_{i,j} = \frac{e^{\mathbf Z_{i,j}}}{\sum_{k=1}^{d} e^{\mathbf Z_{k,j}}};\quad i\in\mathbb N_d,~ j\in\mathbb N_n
\end{align*} 
for $\mathbf{Z}\in \mathbb R^{d\times n}$. Note that each output $\mathbf y_i$ of the self-attention layer directly depends on every input $\mathbf x_i$. This allows the model to learn about the pair-wise relations among all the elements of the input sequence, using a constant number of model parameters that does not depend on the sequence length. Furthermore, compound applications of the self-attention layer result in more complex and improved representations of each element in the sequence, allowing for more complicated relations to be learned. These properties are beneficial for tackling the data association problem in MTT, allowing the model to leverage on complex, long-range patterns in the sequence of measurements when detecting objects.

In practice, Transformers use so-called multi-headed self-attention layers (shown in orange in Fig.\,\ref{fig:encoder-decoder_diagram}), where $\mathbf{X}$ is fed to $n_h$ different self-attention layers (with independent learnable parameters) in parallel, generating $n_h$ different values $\mathbf{Y}_1, \cdots, \mathbf{Y}_{n_h}$. The final output is then computed by vertically stacking the results and applying a linear transformation to reduce the dimensionality:
\begin{align}
    \mathbf{Y} &= \mathbf{W}^0\begin{bmatrix}
        \mathbf{Y}_1 \\
        \vdots       \\
        \mathbf{Y}_{n_h}\end{bmatrix}
\end{align}
where $\mathbf{W}^0\in\mathbb R^{d\times dn_n}$ is a learnable parameter of the multi-head self-attention layer. Finally, $\mathbf{Y}$ is converted back to a sequence $\mathbf{y}_{1:n}=\mathrm{MultiAttention}(\mathbf{x}_{1:n})$. 
These multi-headed self-attention layers are then used to form the two main modules of a Transformer model: the encoder and the decoder.

\subsection{Transformer Encoder}
\label{subsec:transformer_encoder}
The encoder module of the Transformer model is the part responsible for transforming the input sequence into a sequence where each element has now a representation (embedding) that depends on all other elements in the sequence. Specifically, this is done by stacking $N$ ``encoder blocks'' on top of each other, as shown in the left of Fig.\,\ref{fig:encoder-decoder_diagram}. The output for encoder layer $l\in\mathbb N_N$ is computed as 
\begin{align}
    \mathbf t^{(l)}_{1:n} &= \mathrm{MultiAttention}(\mathbf x_{1:n}^{(l-1)})
    \\
    \tilde{\mathbf{t}}^{(l)}_{1:n} &= \mathrm{LayerNorm}(\mathbf x_{1:n}^{(l-1)} + \mathbf t^{(l)}_{1:n})
    \\
    \mathbf x_{1:n}^{(l)} &= \mathrm{LayerNorm}(
    \tilde{\mathbf{t}}^{(l)}_{1:n} + \text{FFN}(\tilde{\mathbf{t}}^{(l)}_{1:n}))~,
\end{align}
where $\mathbf x_{1:n}^{(0)}$ is the input sequence $\mathbf x_{1:n}$, Multi-Attention is a multiheaded self-attention layer, FFN is a fully-connected feed-forward network, and LayerNorm is a Layer Normalization layer \cite{layer_normalization}. The output of the last layer $\mathbf x_{1:n}^{(N)}$ is the encoded input sequence, denoted $\mathbf e_{1:n}$ in Fig.\,\ref{fig:encoder-decoder_diagram}. Once the encoder is trained, each $\mathbf e_i$, $i\in\mathbb N_n$ will contain an improved representation of $\mathbf x_i$ that potentially summarizes the relevant information from all other $\mathbf x_j$, $j\in\mathbb N_n\setminus i$, in a way that is helpful for the task at hand. In MTT, for example, it can contain relevant information about other measurements from the same object at different time steps.

As one can infer from this type of architecture, the encoder blocks are permutation-equivariant, and can be seen as learning mappings between input and output sets. However, in many applications (e.g. translation, sentiment analysis, MTT), there is important information encoded in the order of the elements in the input sequence. To allow the Transformer model to leverage on such information, one usually adds a ``positional encoding'' $\mathbf q_{1:n}$ vector to the input of every encoder (and often decoder) layer, using $\mathbf x^{(l)}_{1:n} + \mathbf q_{1:n}$, in the subsequent computations instead of $\mathbf x_{1:n}$, where $\mathbf q_i = f_p(i)$. The function $f_p$ can either be fixed, usually with sinusoidal components \cite{transformer_paper}, or learnable \cite{DETR}.

\subsection{DETR Decoder}
The Transformer decoder is the module responsible for using the embeddings computed by the encoder to predict the output sequence $\mathbf y_{1:k}$. Different Transformer decoder structures have been proposed for different problems \cite{transformer_paper, evolved_transformer, decoder_for_speech_recon_and_translation}, and the one used in this paper is similar to the one proposed in \cite{DETR}. Specifically, instead of performing predictions autoregressively as in \cite{transformer_paper}, the DEtection TRansformer (DETR) \cite{DETR} decoder takes as input the encoded input sequence and a sequence of ``object queries''  $\mathbf o_{1:k}$ (learnable vectors), and computes all of its predictions in parallel. Each object query $\mathbf o_i$ learns to attend to aspects of the embeddings $\mathbf e_{1:n}$ which are helpful for predicting the output $\mathbf y_i$. In MTT, for example, an object query $\mathbf o_i$ can potentially learn to only attend to the embeddings of the true measurements from a certain object, making it possible for the decoder self-attention layers to perform the easier task of regression.

As shown in Fig\, \ref{fig:encoder-decoder_diagram}, the decoder is comprised of $M$ ``decoder blocks'' stacked on top of each other, where the input of the next block is the output of the previous one. Similarly to the encoder, the input is processed by layers of self-attention followed by an FFN, with normalization layers in-between. The Cross-attention layer shown in Fig\, \ref{fig:encoder-decoder_diagram} is a regular multi-head attention layer as described in section \ref{sec:self-attention}, with the difference that the matrices $\mathbf K$ and $\mathbf V$ in \eqref{eq:self-attention_qkv} are respectively computed by multiplying $\mathbf W_K$ and $\mathbf W_V$ with $\mathbf e_{1:n}$, while $\mathbf Q$ is computed as usual, by multiplying $\mathbf W_Q$ with the output from the previous normalization layer. All of the subsequent computations are the same as regular multiheaded self-attention. 

\section{MTT using Transformers}
\label{sec:mtt_using_transformers}
This section describes our proposed transformer architecture MT3: MultiTarget Tracking with Transformers. We approach the task of model-based MTT using deep learning, where the available transition and observation models are used for generating unlimited training data. In comparison to traditional MTT algorithms, we sidestep the need for using recursive pruning strategies that impair estimation performance by proposing a network structure able to implicitly reason about the data association hypotheses across the entire sequence, given all measurements.


\subsection{MT3 Architecture}
\label{subsec:high_level_mt3}
\begin{figure}
    \centering

\tikzset{every picture/.style={line width=0.75pt}} 

\begin{tikzpicture}[x=0.75pt,y=0.75pt,yscale=-1,xscale=1]

\draw  [fill={rgb, 255:red, 209; green, 255; blue, 195 }  ,fill opacity=1 ][dash pattern={on 4.5pt off 4.5pt}] (135,44.57) -- (310.4,44.57) -- (310.4,170.49) -- (135,170.49) -- cycle ;
\draw    (131.5,251) -- (131.5,238.13) -- (131.5,224) ;
\draw [shift={(131.5,221)}, rotate = 450] [fill={rgb, 255:red, 0; green, 0; blue, 0 }  ][line width=0.08]  [draw opacity=0] (8.93,-4.29) -- (0,0) -- (8.93,4.29) -- (5.93,0) -- cycle    ;
\draw    (161,202.6) -- (298,202.6) ;
\draw [shift={(301,202.6)}, rotate = 180] [fill={rgb, 255:red, 0; green, 0; blue, 0 }  ][line width=0.08]  [draw opacity=0] (8.93,-4.29) -- (0,0) -- (8.93,4.29) -- (5.93,0) -- cycle    ;
\draw    (315,220) -- (315,247) ;
\draw [shift={(315,250)}, rotate = 270] [fill={rgb, 255:red, 0; green, 0; blue, 0 }  ][line width=0.08]  [draw opacity=0] (8.93,-4.29) -- (0,0) -- (8.93,4.29) -- (5.93,0) -- cycle    ;
\draw    (289,72.56) -- (333.4,72.56) -- (333.4,183.57) ;
\draw [shift={(333.4,186.57)}, rotate = 270] [fill={rgb, 255:red, 0; green, 0; blue, 0 }  ][line width=0.08]  [draw opacity=0] (8.93,-4.29) -- (0,0) -- (8.93,4.29) -- (5.93,0) -- cycle    ;
\draw  [fill={rgb, 255:red, 217; green, 217; blue, 217 }  ,fill opacity=1 ] (99.5,194.1) .. controls (99.5,190.45) and (102.45,187.5) .. (106.1,187.5) -- (154.4,187.5) .. controls (158.05,187.5) and (161,190.45) .. (161,194.1) -- (161,213.9) .. controls (161,217.55) and (158.05,220.5) .. (154.4,220.5) -- (106.1,220.5) .. controls (102.45,220.5) and (99.5,217.55) .. (99.5,213.9) -- cycle ;
\draw  [fill={rgb, 255:red, 217; green, 217; blue, 217 }  ,fill opacity=1 ] (301.5,193.1) .. controls (301.5,189.45) and (304.45,186.5) .. (308.1,186.5) -- (356.4,186.5) .. controls (360.05,186.5) and (363,189.45) .. (363,193.1) -- (363,212.9) .. controls (363,216.55) and (360.05,219.5) .. (356.4,219.5) -- (308.1,219.5) .. controls (304.45,219.5) and (301.5,216.55) .. (301.5,212.9) -- cycle ;
\draw  [fill={rgb, 255:red, 177; green, 245; blue, 249 }  ,fill opacity=1 ] (154.67,117.94) .. controls (154.67,115.5) and (156.64,113.53) .. (159.08,113.53) -- (191.39,113.53) .. controls (193.82,113.53) and (195.8,115.5) .. (195.8,117.94) -- (195.8,131.19) .. controls (195.8,133.62) and (193.82,135.6) .. (191.39,135.6) -- (159.08,135.6) .. controls (156.64,135.6) and (154.67,133.62) .. (154.67,131.19) -- cycle ;

\draw  [fill={rgb, 255:red, 177; green, 245; blue, 249 }  ,fill opacity=1 ] (250.67,117.34) .. controls (250.67,114.9) and (252.64,112.93) .. (255.08,112.93) -- (287.39,112.93) .. controls (289.82,112.93) and (291.8,114.9) .. (291.8,117.34) -- (291.8,130.59) .. controls (291.8,133.02) and (289.82,135) .. (287.39,135) -- (255.08,135) .. controls (252.64,135) and (250.67,133.02) .. (250.67,130.59) -- cycle ;

\draw  [fill={rgb, 255:red, 255; green, 149; blue, 151 }  ,fill opacity=1 ] (252.13,64.44) .. controls (252.13,62) and (254.11,60.03) .. (256.55,60.03) -- (285.05,60.03) .. controls (287.49,60.03) and (289.47,62) .. (289.47,64.44) -- (289.47,77.69) .. controls (289.47,80.12) and (287.49,82.1) .. (285.05,82.1) -- (256.55,82.1) .. controls (254.11,82.1) and (252.13,80.12) .. (252.13,77.69) -- cycle ;

\draw    (221,157.57) -- (272,157.57) -- (272,138.57) ;
\draw [shift={(272,135.57)}, rotate = 450] [fill={rgb, 255:red, 0; green, 0; blue, 0 }  ][line width=0.08]  [draw opacity=0] (8.93,-4.29) -- (0,0) -- (8.93,4.29) -- (5.93,0) -- cycle    ;
\draw    (201,70.63) -- (248.2,70.63) ;
\draw [shift={(251.2,70.63)}, rotate = 180] [fill={rgb, 255:red, 0; green, 0; blue, 0 }  ][line width=0.08]  [draw opacity=0] (8.93,-4.29) -- (0,0) -- (8.93,4.29) -- (5.93,0) -- cycle    ;
\draw    (224,157.57) -- (224,203.09) ;
\draw    (221,157.57) -- (176.2,157.57) -- (176.2,139.17) ;
\draw [shift={(176.2,136.17)}, rotate = 450] [fill={rgb, 255:red, 0; green, 0; blue, 0 }  ][line width=0.08]  [draw opacity=0] (8.93,-4.29) -- (0,0) -- (8.93,4.29) -- (5.93,0) -- cycle    ;
\draw    (270.4,112.17) -- (270.4,85.57) ;
\draw [shift={(270.4,82.57)}, rotate = 450] [fill={rgb, 255:red, 0; green, 0; blue, 0 }  ][line width=0.08]  [draw opacity=0] (8.93,-4.29) -- (0,0) -- (8.93,4.29) -- (5.93,0) -- cycle    ;
\draw    (175.8,113.77) -- (175.8,84.77) ;
\draw [shift={(175.8,81.77)}, rotate = 450] [fill={rgb, 255:red, 0; green, 0; blue, 0 }  ][line width=0.08]  [draw opacity=0] (8.93,-4.29) -- (0,0) -- (8.93,4.29) -- (5.93,0) -- cycle    ;
\draw  [fill={rgb, 255:red, 255; green, 255; blue, 255 }  ,fill opacity=1 ] (150.53,63.64) .. controls (150.53,61.2) and (152.51,59.23) .. (154.95,59.23) -- (196.39,59.23) .. controls (198.82,59.23) and (200.8,61.2) .. (200.8,63.64) -- (200.8,76.89) .. controls (200.8,79.32) and (198.82,81.3) .. (196.39,81.3) -- (154.95,81.3) .. controls (152.51,81.3) and (150.53,79.32) .. (150.53,76.89) -- cycle ;

\draw    (270,28.71) -- (270,56.71) ;
\draw [shift={(270,59.71)}, rotate = 270] [fill={rgb, 255:red, 0; green, 0; blue, 0 }  ][line width=0.08]  [draw opacity=0] (8.93,-4.29) -- (0,0) -- (8.93,4.29) -- (5.93,0) -- cycle    ;
\draw    (351,219.8) -- (351,246.8) ;
\draw [shift={(351,249.8)}, rotate = 270] [fill={rgb, 255:red, 0; green, 0; blue, 0 }  ][line width=0.08]  [draw opacity=0] (8.93,-4.29) -- (0,0) -- (8.93,4.29) -- (5.93,0) -- cycle    ;

\draw (122,253.25) node [anchor=north west][inner sep=0.75pt]    {$\mathbf{z}_{1:n}$};
\draw (321.8,56) node [anchor=north west][inner sep=0.75pt]    {$\mathbf{o}_{1:k}$};
\draw (219.4,205) node [anchor=north west][inner sep=0.75pt]    {$\mathbf{e}_{1:n}$};
\draw (313,196.5) node [anchor=north west][inner sep=0.75pt]  [font=\footnotesize] [align=left] {Decoder};
\draw (111,197.5) node [anchor=north west][inner sep=0.75pt]  [font=\footnotesize] [align=left] {Encoder};
\draw (134.07,29) node [anchor=north west][inner sep=0.75pt]  [font=\footnotesize] [align=left] {Selection mechanism};
\draw (212.33,59) node [anchor=north west][inner sep=0.75pt]  [font=\scriptsize]  {$m_{1:n}$};
\draw (248.74,94.03) node [anchor=north west][inner sep=0.75pt]  [font=\scriptsize]  {$\boldsymbol{\delta}_{1:n}$};
\draw (256,64.44) node [anchor=north west][inner sep=0.75pt]  [font=\footnotesize] [align=left] {Top-k};
\draw (259.5,118.5) node [anchor=north west][inner sep=0.75pt]  [font=\footnotesize] [align=left] {FFN};
\draw (163.5,119) node [anchor=north west][inner sep=0.75pt]  [font=\footnotesize] [align=left] {FFN};
\draw (155,64.64) node [anchor=north west][inner sep=0.75pt]  [font=\footnotesize] [align=left] {Softmax};
\draw (258,15) node [anchor=north west][inner sep=0.75pt]  [font=\normalsize]  {$\mathbf{z}_{1:n}$};
\draw (344,251.2) node [anchor=north west][inner sep=0.75pt]    {$\hat{\mathbf{x}}^{T}_{1:k}$};
\draw (302,255.2) node [anchor=north west][inner sep=0.75pt]    {$p_{1:k}$};

\end{tikzpicture}

    \caption{High-level diagram of the MT3 architecture. The measurements $\mathbf z_{1:n}$ are processed by the encoder, producing embeddings $\mathbf e_{1:n}$. These embeddings are used by a selection mechanism for generating object queries $\mathbf o_{1:k}$. Lastly, $\mathbf e_{1:n}$ and $\mathbf o_{1:k}$ are used by the decoder for predicting the object states, $\hat{\mathbf x}_{1:k}^T$, and corresponding existence probabilities $p_{1:k}$.}
    \label{fig:mt3_diagram}
\end{figure}
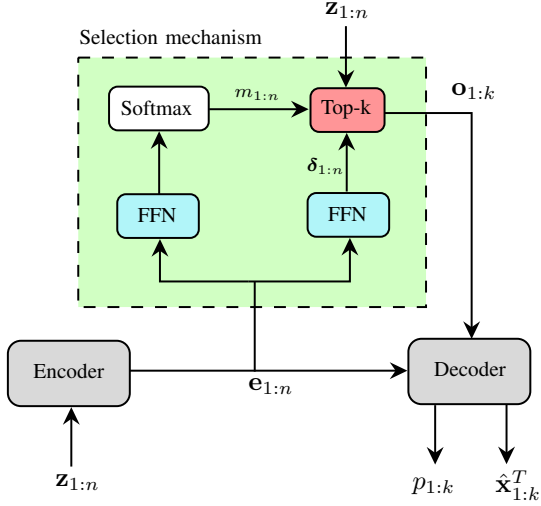

MT3 uses an encoder-decoder architecture for processing the input measurement sequence, as shown in Fig.\,\ref{fig:mt3_diagram}. The idea is that the Transformer encoder can process the measurement sequence, generating a new, improved representation for each measurement capable of encoding helpful properties such as which measurements come from the same object, which are clutter measurements, etc. We feed these new representations into a modified DETR decoder, which can then leverage on them for implicitly performing soft data associations using cross-attention, and subsequent state estimation with the decoder self-attention.

Concretely, the measurements $[\mathbb Z^{T-\tau},\cdots,\mathbb Z^T]$ are collected in a sequence
\begin{equation}
    \label{eq:sequence_of_measurements_definition}
    \mathbf z_{1:n} = [\mathbf z^{T-\tau}_1, \cdots, \mathbf z^{T-\tau}_{n^{T-\tau}}, \cdots, \mathbf z^T_1, \cdots, \mathbf z^T_{n^T}],
\end{equation}
where $n^t\doteq |\mathbb Z^t|$, $n=\sum_{i=T-\tau}^T n^i_z$, and the elements of each $\mathbb Z^i$ are added in random order to the sequence. 
As shown in Fig.\, \ref{fig:mt3_diagram}, the sequence $\mathbf z_{1:n}$ is fed to a Transformer encoder ($N$ blocks), which generates embeddings for each measurement, resulting in a sequence $\mathbf e_{1:n}$. The sequence of embeddings ${\bf e}_{1:n}$ is then fed to a Transformer decoder (containing $M$ encoder blockes), together with object queries $\mathbf o_{1:k}$ produced by a selection mechanism (explained in Section \ref{subsec:selection_mechanism}), to produce the output: estimated states for the objects present at time-step $T$, $\hat{\mathbf x}_{1:k}^T$, and corresponding existence probabilities, $p_{1:k}$.

\subsection{Selection Mechanism}
\label{subsec:selection_mechanism}
In order to produce the object queries $\mathbf o_{1:k}$, a selection mechanism similar to the two-stage encoder proposed in \cite{deformable-DETR} is used: instead of using learned object queries that are the same for all sequences $\mathbf z_{1:n}$, the $k$ most promising measurements in $\mathbf z_{1:n}$ are used as starting points, in order to facilitate the task of the decoder. To that end, each embedding $\mathbf e_i$ is fed to two different FFN layers, see Fig.\, \ref{fig:mt3_diagram}. To select the most promising measurements, scores $m_i=\text{Softmax}(\text{FFN}_1(\mathbf e_i))\in[0,1],~i\in\mathbb N_n$ are computed, and the measurements with the top-k-scores are then selected as candidates for the object queries. In order to allow more flexibility, the second FFN computes offsets $\boldsymbol\delta_i=\text{FFN}_2(\mathbf e_i)\in\mathbb R^{d_z},~i\in\mathbb N_n$, which are added to the chosen measurements:
\begin{equation}
    \label{eq:z_tildes}
    \tilde{\mathbf{z}}_i = \mathbf z_{r_i}+\boldsymbol\delta_{r_i},\quad i\in\mathbb N_k~.
\end{equation}
where $\mathbf{r}=\text{argsort}(m_{1:n})$, and $\text{argsort}$ is a function that returns the indices that would sort the input array (i.e., $m_{r_i}\ge m_{r_j} \iff i>j$, for $r=\text{argsort}(m_{1:n})$). The sequence $\tilde{\bf z}_{1:k}$ is then fed to an FFN layer, producing the object queries $\mathbf o_{1:k}$ for the decoder
\begin{equation}
    \label{eq:object_queries}
    \mathbf o_i = \text{FFN}(\tilde{\mathbf z}_i),\quad i\in\mathbb N_k~,
\end{equation}
where each element $\mathbf o_i \in \mathbb R^{d'}$, and $d'>d_z$ is a hyperparameter.

\subsection{Iterative Refinement of State}
\label{subsec:iterative_refinement}

We also adopt the idea of iterative refinement \cite{iterative_refinement_1, iterative_refinement_2, iterative_refinement_3, deformable-DETR}. Instead of directly predicting the quantity of interest, each layer in the decoder predicts adjustments to the predictions made in the previous layer. In MT3, the output $\mathbf y_{1:k}^l$ from each decoder layer $l\in\mathbb N_M$ is fed through an FFN, producing adjustments $\mathbf \Delta^l_{1:k}$ for each of the $k$ objects. The estimated position at layer $L$ is then computed as
\begin{equation}
    \label{eq:improved_measurements}
    \hat{\mathbf{ x}}^{T, L}_{i} = \tilde{\mathbf z}_i + \sum_{l=1}^{L} \mathbf \Delta_i^l~,\quad i\in \mathbb N_k~.
\end{equation}
The existence probabilities, however, are not iteratively refined, and are computed directly by feeding the output of each decoder layer $l$ through an FFN: 
\begin{equation}
    \label{eq:existence_probabilities}
    p_i^l=\text{FFN}(\mathbf y_i^l),\quad i\in\mathbb N_k~.
\end{equation}
Therefore, at each decoder layer $l$, estimates $\hat{\mathbf{ x}}^{T, l}_{1:k}$ and existence probabilities $p_{1:k}^l$ are produced. The final output of MT3, shown in Fig.\,\ref{fig:mt3_diagram}, is defined as the estimates computed at the last decoder layer $M$: $\hat{\mathbf x}_{1:k}^T = \hat{\mathbf x}_{1:k}^{T,M}$ and $p_{1:k}=p_{1:k}^M$.

\subsection{Loss Function}
\label{subsec:loss_function}
We use the same loss function as defined in \cite{DETR}, which is based on the localization error and probabilities of missed and false targets, given the best match between ground-truth and predictions. Instead of applying such loss only to the final outputs $(\hat{\mathbf x}_{1:k}^T, p_{1:k})$, we apply it to the estimates produced at all the decoder layers, summing all contributions together. This was shown to improve performance for deep Transformer decoder structures, accelerating learning \cite{auxiliary_decoding_losses, DETR}.

Given a prediction made by the Transformer at decoder layer $l$:
$$ \mathbf a_{1:k}^l = \left((\hat{\mathbf{ x}}^{T,l}_1, p_1^l), ..., (\hat{\mathbf{ x}}^{T,l}_k, p_k^l)\right)~, $$
and ground-truth states for all objects present at time-step $T$:
$$ \mathbf g_{1:|\mathbb X^T|} = \left(\mathbf x^T_1, ..., \mathbf x^T_{|\mathbb X^T|}\right)~, $$
we assume\footnote{We choose a value of $k$ which is large in comparison to the generative model, and enforce the inequality constraint during training by not adding more than $k$ objects in a scene, $k$ being the number of object queries of the model being trained. This restriction is only applied during training, during inference this loss needs not be computed.}, in the same fashion as \cite{DETR}, that $k\geq |\mathbb X^T|$ and then pad $\mathbf g_{1:|\mathbb X^T|}$ with $\varnothing$ elements so that $k=|\mathbb X^T|$. The best match is then computed as
\begin{equation}
    \label{eq:optimal_permutation_definition}
    \sigma^l_* = \arg\min_\sigma \sum_{i=1}^k \mathcal L_m(\mathbf a_i^l, \mathbf g_{\sigma(i)})
\end{equation}
where $\sigma$ is a permutation function:
\begin{align*}
    &\sigma: \mathbb N_k \rightarrow \mathbb N_k ~|~ \sigma(i) = \sigma(j) \Rightarrow i = j~,
\end{align*}
and $\mathcal L_m$ is the matching loss, defined as
\begin{equation}
    \mathcal L_m(\mathbf a_i^l, \mathbf g_j) = 
    \begin{cases}
        0 & \text{ if } \mathbf a_j^l=\varnothing 
        \\ 
        \Vert\hat{\mathbf{ x}}^{T,l}_i - \mathbf x^T_i\Vert - p_i^l & \text{ otherwise. } 
    \end{cases}
\end{equation}
The optimization problem in \eqref{eq:optimal_permutation_definition} can be efficiently solved using the Hungarian algorithm \cite{hungarian-method}. Given the best match $\sigma^l_*$ at each layer $l$, the final loss used to train the model is:
\begin{equation}
    \label{eq:total_loss}
    \mathcal L_T(\mathbf a_{1:k}^{1:M}, \mathbf g_{1:k})=\sum_{l=1}^M\sum_{i=1}^k\mathcal L(\mathbf a_i^l, \mathbf g_{\sigma^l_*(i)})
\end{equation}
where $\mathcal L$ is defined as
\begin{equation}
    \mathcal L(\mathbf a_i^l, \mathbf g_j) = 
    \begin{cases}
        -\log(1-p_i) & \text{ if } \mathbf g_j= \varnothing
        \\ 
        \Vert \hat{\mathbf{ x}}^{T,l}_i - \mathbf x^T_i \Vert-\log(p_i) & \text{ otherwise.}  
\end{cases}
\end{equation}

\subsection{Contrastive Auxiliary Learning}
\label{subsec:contrastive_learning}
Training on auxiliary tasks often helps the training process and the generalization performance of the final model, by leveraging on information helpful for the task which is present in the inputs, but not being directly used by the primary loss chosen for training \cite{multi_task_learning}. In MTT, this could be the information of whether a measurement is clutter or not, or if it comes from the same object as another measurement. Although such information is helpful for solving the task, it is not directly used by the loss chosen in Section \ref{subsec:loss_function}. In order to steer the model into quickly learning how to perform this task, we use an idea inspired by Supervised Contrastive Learning \cite{SCL} to create the contrastive loss $\mathcal L_c$, with the aid of object labels for each of the measurements. The idea is to encourage the encoder to generate embeddings of measurements which are similar for measurements coming from the same object, but dissimilar for measurements coming from different objects. False measurements are treated as coming all from the same dummy object (and therefore their encoded representations should be similar to each other, but dissimilar to the representations of all other measurements).

Concretely, for a given sequence of measurements $\mathbf z_{1:n}$, let $b_i$ be the object from which measurement $\mathbf z_i$ came from, $i\in\mathbb N_n$. We define $\mathbb P_i = \{j \in \mathbb N_n ~|~ j\neq i~,~b_i = b_j\}$,
that is, $\mathbb P_i$ is the set of indices of measurements that came from the same object as $\mathbf z_i$. The auxiliary loss $\mathcal L_c$ is then defined as:
\begin{equation}
    \label{eq:contrastive_loss}
    \mathcal L_c(\mathbf u_{1:n}, \mathbb P_{1:n}) = \alpha\sum_{i=1}^n\frac{-1}{|\mathbb P_i|}\sum_{i^+\in\mathbb P_i}\log \frac{e^{\mathbf u_i^\top \mathbf u_{i^+}}}{\sum\limits_{j\in \mathbb N_n\setminus i}e^{\mathbf u_i^\top \mathbf u_j}}
\end{equation}
where, $\mathbf u_{1:n}$ is a sequence of vectors computed by processing each $\mathbf e_i, i\in\mathbb N_n$ (embeddings of $\mathbf z_{1:n}$ computed by the encoder) with an FFN layer and normalizing to unit length, and $\alpha\geq 0$ is a hyperparameter used to control the importance of this task. Minimizing this loss can be understood as making $\mathbf u_i^\top\mathbf u_j$ large when $\mathbf z_i$ and $\mathbf z_j$ are from the same object, but small when they are from different objects. Training on the sum of this auxiliary loss and the loss defined in \eqref{eq:total_loss} accelerated learning and improved the final performance of the trained model in both tasks considered in Section \ref{sec:results}.

\subsection{Preprocessing}
\label{subsec:preprocessing}
We perform three steps of preprocessing: normalization, dimensionality augmentation, and addition of a temporal encoding, whose main aspects are explained in this subsection. For additional information we refer the reader to our code.
For preprocessing, we first normalize all dimensions of each measurement vector $\mathbf z_i$ in $\mathbf z_{1:n}$ to be in $[0,1]$, using the known field-of-view dimensions. Then, we increase the dimension of each measurement vector using a linear transformation 
\begin{equation}
    \mathbf z'_i = \mathbf{W} \mathbf z_i + \mathbf b, \quad i\in\mathbb N_n~,
\end{equation}
where $\mathbf z'_i\in\mathbb R^{d'}$ is the dimensionality augmented measurement vector, $d'$ was defined in Section \ref{subsec:selection_mechanism}, and $\mathbf{W}\in\mathbb R^{d'\times d_z}$, $\mathbf b\in\mathbb R^{d'}$ are learnable parameters. Lastly, we add positional encodings to the inputs of all the encoder and decoder layers, with learnable $f_p$ as described in section \ref{subsec:transformer_encoder}, similar to \cite{DETR}. However, instead of using the position of the elements as input to $f_p$, we use the corresponding time-step for that position in the sequence. The intuition is that this is the relevant quantity for the problem at hand, not the position within the sequence, since different positions can map to the same time index, as shown in \eqref{eq:sequence_of_measurements_definition}.

\section{Results}
\label{sec:results}
We evaluate the performance of MT3 in two tasks of different complexity using synthetic data, and compare it against two state-of-the-art Bayesian RFS filters: the PMBM filter \cite{pmbm_older, PMBM}, and the $\delta$-GLMB filter \cite{GLMB}, both with linear/Gaussian implementation. Furthermore, we perform a series of ablation studies to evaluate the importance of different components of the MT3 architecture.

\subsection{Definition of the Tasks}
We compare the performance of the three trackers under two different variations, henceforth referred to as Task 1 and Task 2. For both tasks, $\tau=20$, $\lambda_b(\mathbf x)=10^{-3}$, $p_s(\mathbf x)=0.95$, the field of view is the 2D square $[-10, 10]\times[-10, 10]$, and we use Poisson models with parameter $\lambda_0$ for the initial
number of objects. The motion model used is the nearly constant velocity model, defined as:
\begin{equation}
    f(\mathbf x^{t+1}|\mathbf x^t) = 
    \mathcal N\left(\begin{bmatrix}
        \mathbf I & \mathbf I\Delta_t
        \\ 
        \mathbf 0 & \mathbf I
    \end{bmatrix} \mathbf x^t~,~\sigma_q^2
    \begin{bmatrix}
        \mathbf I\frac{\Delta_t^3}{3} & \mathbf I\frac{\Delta_t^2}{2}
        \\
        \mathbf I\frac{\Delta_t^2}{2} & \mathbf I\Delta_t
    \end{bmatrix}\right)
\end{equation}
where $\mathbf x^{t+1}, \mathbf x^t\in \mathbb R^{d_x}$, $d_x=4$ represents target position and velocity in 2D,  $\Delta_t=0.1$ is the sampling period, $\sigma_q$ controls the magnitude of the process noise, and the state for newborn objects is sampled from $\mathcal N(\mathbf 0, 3\mathbf I)$.
The linear Gaussian measurement model is used with measurement likelihood
\begin{equation}
    g\left(\mathbf z^t~|~\mathbf x^t)=  \mathcal N(\mathbf H \mathbf x^t, \mathbf I\sigma_z^2\right)
\end{equation}
where $\bf H$ selects the position components from $\mathbf x^t$, and $\sigma_z$ controls the magnitude of the measurement noise.

Task 1 has $\lambda_0=4$, $p_d=0.9$, $\lambda_c=0.05$, $\sigma_q=0.5$, and $\sigma_z=0.1$. We expect the model-based Bayesian trackers to be able to approximate well the optimal solution in this context, making them a strong benchmark for evaluating the Transformer-based model. Task 2 has  $\lambda_0=6$, $p_d=0.8$, $\lambda_c=0.075$, $\sigma_q=0.9$, and $\sigma_z=0.3$. The lower signal-to-noise ratio in this task makes it considerably harder for conventional MTT algorithms to perform well with a feasible computational complexity, since the number of probable hypothesis to keep track of grows considerably.

\subsection{Algorithms}

\subsubsection{PMBM}
The PMBM filter provides a closed-form solution for multiple point object tracking with standard multitarget models with Poisson birth. The unknown data associations lead to an intractably large number of terms in the PMBM density. To achieve computational tractability of the PMBM filter, it is necessary to reduce the number of PMBM parameters. First, gating is performed to remove unlikely measurement-to-object associations, and the gating size is 20. Second, we use Murty’s algorithm \cite{murty} to find up to 200 best assignments. Third, we prune multi-Bernoullis with weight smaller than $10^{-4}$, Bernoulli components with probability of existence smaller than $10^{-5}$ and Poisson mixture components with weight smaller than $10^{-5}$. Object state estimation is performed by obtaining the global hypothesis with highest weight at time-step $T$ and reporting the means of the Bernoullis whose existence is above 0.5 (Estimator 1 in \cite{PMBM}).

\subsubsection{$\delta$-GLMB}
The $\delta$-GLMB filter provides a closed-form solution for MTT when the object birth model is a multi-Bernoulli (mixture). To achieve computational tractability of the $\delta$-GLMB filter, gating and Murty's algorithm are implemented to remove unlikely measurement-to-object associations as in the PMBM filter. In addition, we prune multi-Bernoullis with weight smaller than $10^{-4}$. Object states are extracted from the global hypothesis with the highest weight.

\subsubsection{MT3}
The architecture choice used for the encoder is the same as in \cite{transformer_paper}, and for decoder the same as in \cite{DETR}. We use 6 encoder and 6 decoder layers, and in all of them the multiheaded self-attention layers have 8 heads. All FFN layers in the encoder and decoder layers are comprised of 2048 hidden units, and we use a dropout rate of 0.1 during training. The increased state dimensionality $d'$ is set to $256$. The FFNs used for predicting measurement adjustments $\boldsymbol \delta_i$ in \eqref{eq:z_tildes}, state refinements $\mathbf \Delta_{1:k}^l$ in \eqref{eq:improved_measurements}, object queries $\mathbf o_i$ in \eqref{eq:object_queries}, and existence probabilities $p_i^l$ in \eqref{eq:existence_probabilities} are all comprised of single-layer neural networks with 128 hidden units, while the FFN used in the computation of $\mathbf u_i$'s in \eqref{eq:contrastive_loss} has 256. Furthermore, in all our experiments we set the contrastive loss weight $\alpha=4.0$, and we extract target estimates from $\hat{{\bf x}}_i^t$ with $p_i \geq 0.9$. MT3 was trained for each task starting with random weights, for 800k and 600k gradient descent steps for Task 1 and Task 2, respectively, using ADAM optimizer \cite{ADAM} with a batch size of 32. Each batch of samples is generated using the known transition and observation models available for each task. The initial learning rate was set to $5\cdot10^{-5}$, and whenever the total loss value $\mathcal L_T$ did not decrease for 50k consecutive time-steps, the learning rate was reduced by a factor of 4. The code used to define, train, and evaluate our model is made publicly available.

\subsection{Performance Metrics}
We evaluate the performance of each algorithm using the Generalized optimal sub-pattern assignment (GOSPA) metric \cite{GOSPA} with $\alpha=2$ between predictions $\mathbb{\hat{X}}=\{\hat{\mathbf{ x}}_1^T,\cdots,\hat{\mathbf{ x}}_{|\hat{\mathbb X}|}^T\}$ and the ground-truth states $\mathbb X=\{\mathbf x_1^T,\cdots,\mathbf x_{|\mathbb X|}^T\}$, defined as
\begin{equation}
\label{eq:gospa}
    \begin{aligned}
        &d_p^{(c, 2)}(\mathbb{\hat X}, \mathbb X)=
        \\
        &\min_{\gamma\in\Gamma}\Bigg( {\underbrace{\sum_{(i, j)\in\gamma} d(\hat{\mathbf{ x}}_i^T, \mathbf x_j^T)^p}_{\text{Localization}}} + {\underbrace{\frac{c^p}{2}(|\mathbb{\hat X}|-|\gamma|)}_\text{False detections}} + {\underbrace{\frac{c^p}{2}(|\mathbb X|-|\gamma|)}_\text{Missed detections}} \Bigg)^{\frac{1}{p}}
    \end{aligned}
\end{equation}
where the minimization is over assignment sets between the elements of $X$ and $Y$, such that $\gamma\subseteq\{1,\cdots,|\mathbb{\hat X}|\}\times\{1,\cdots, |\mathbb X|\}$, while $(i,j), (i,j')\in\gamma\implies j=j'$, and $(i,j), (i',j)\in\gamma\implies i=i'$. In all our experiments we use $c=2.0$, $p=1$, and
\begin{equation}
    d(\hat{\mathbf{ x}}_i^T, \mathbf x_i^T) = \Vert\mathbf H \hat{\mathbf x}_i^T - \mathbf H \mathbf x_i^T \Vert~.
\end{equation}
We evaluate the estimation performance of the three trackers using Monte Carlo simulation with 1000 runs. 

\subsection{Task 1 Results}
The resulting average GOSPA scores for Task 1 are shown in Table \ref{tab:gospa_scores_easy_task}, along with corresponding decompositions. The localization decomposition is provided after normalizing it by the number of detected objects, so as to provide an easier comparison between methods with large differences in missed detection rates.d In terms of total GOSPA error, PMBM performs best in this task, with MT3 achieving a close second place. The $\delta$-GLMB filter attains worse performance than PMBM in this task, mainly because PMBM has a more efficient hypothesis structure than $\delta$-GLMB \cite{angel_paper}. MT3 obtains the lowest missed and false detection rates, which implies that it performs the data association task the best.
\begin{table}[b]
    \centering
    \caption{GOSPA and its decompositions for Task 1.  \label{tab:gospa_scores_easy_task}}
   
    \begin{tabular}{@{}lllll@{}}
    \toprule
    \textbf{Method}  & \textbf{GOSPA} & \textbf{\begin{tabular}[c]{@{}l@{}}Localization\end{tabular}} & \textbf{\begin{tabular}[c]{@{}l@{}}Missed\end{tabular}} & \textbf{\begin{tabular}[c]{@{}l@{}}False\end{tabular}} \\ \midrule
    PMBM         & \textbf{1.267} & 0.102 & 0.632 & 0.195 \\ 
    $\delta$-GLMB         & 1.863 & \textbf{0.098} & 1.137 & 0.335 \\
    MT3          & 1.277 & 0.141 & \textbf{0.528} & \textbf{0.094} \\
    \bottomrule
    \end{tabular}
\end{table}

\begin{figure}[t]
    \centering
    \includegraphics[width=0.48\textwidth]{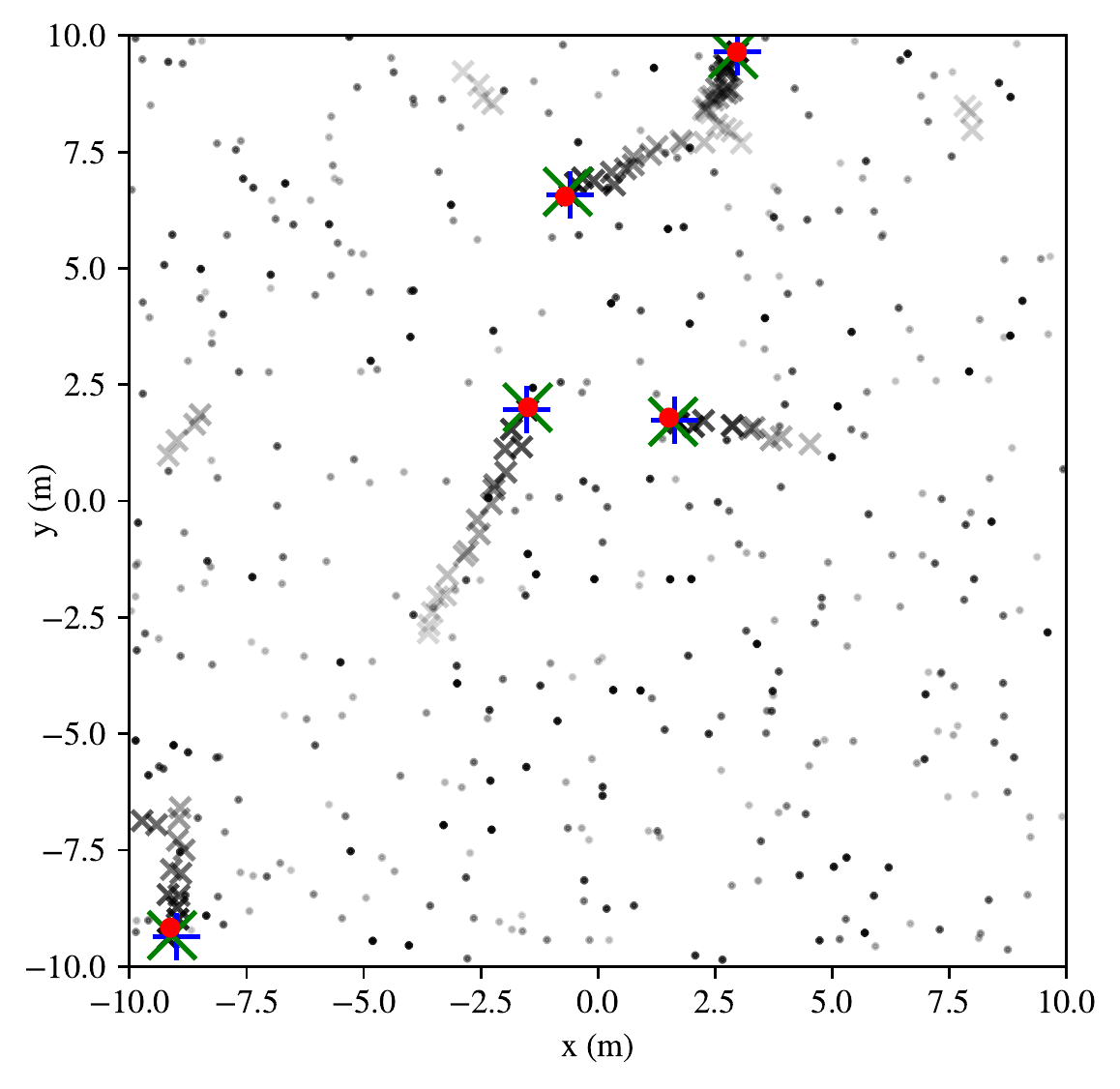}
    \caption{Evaluation sample for Task 1. The measurements generated during the $\tau=20$ simulation time-steps are shown in black, with their transparency illustrating the time-step in which they were taken (more opaque x's correspond to more recent measurements, closer to $t=T$). Clutter is depicted as small black circles and true measurements as black crosses. The ground-truth object positions are shown as red circles, along with the predictions for the two best methods: PMBM as green crosses and MT3 as blue +'s.}
    \label{fig:simple_task_visualization}
\end{figure}

As a simple visual assessment of the task complexity as well as the trackers' performance, an evaluation sample from this task is plotted in Fig. \ref{fig:simple_task_visualization}, where we show the predictions for the best and second-best performing trackers.
From the figure we can see that neither PMBM nor MT3 had difficulty in finding all targets among the clutter. Note that the faint black crosses in the borders of the image correspond to measurements generated by objects which left the scene at a time-step prior to $t=T$, so no predictions are made for them.

\subsection{Task 2 Results}
\begin{figure}[t]
    \centering
    \includegraphics[width=0.48\textwidth]{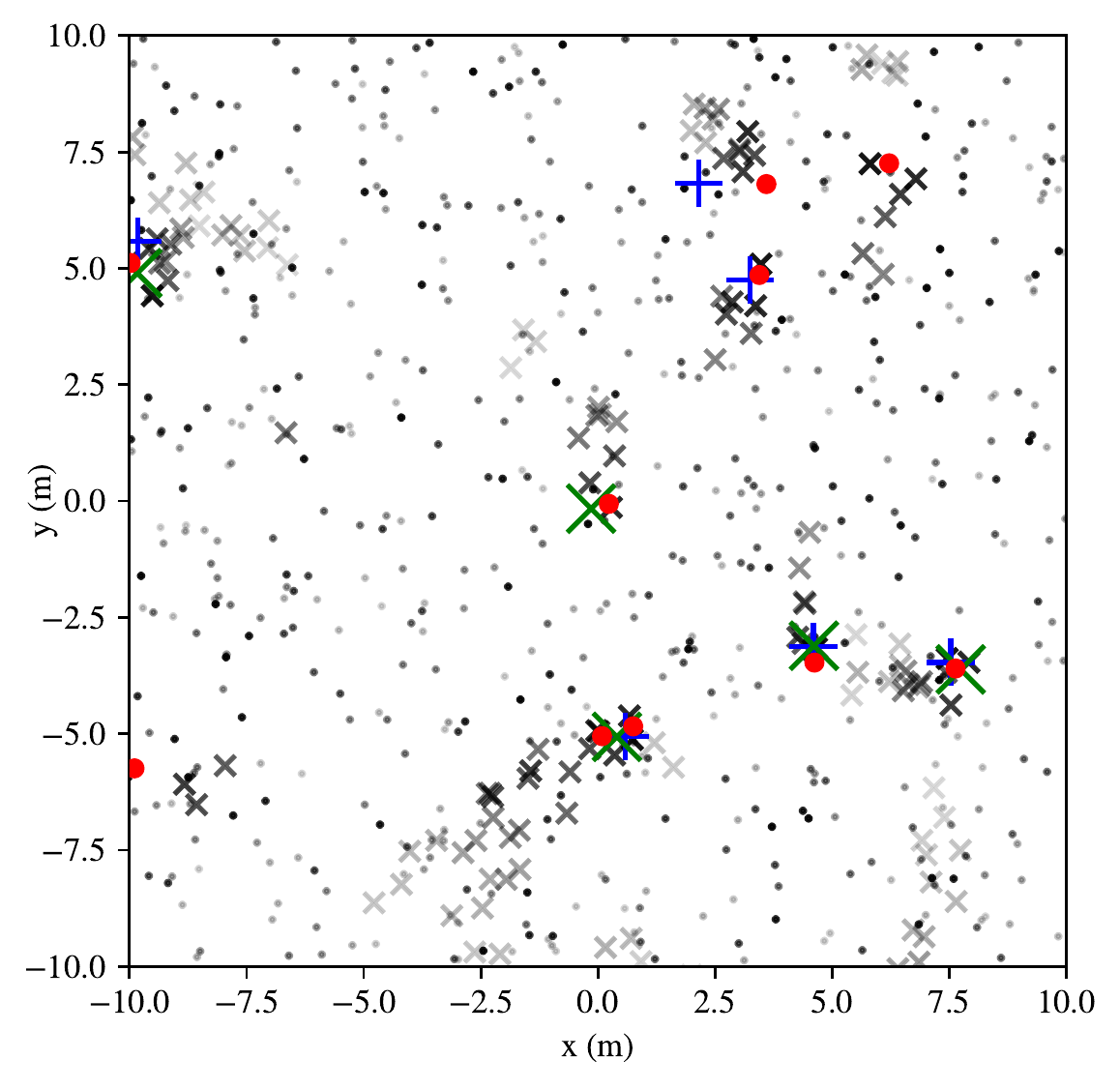}
    \caption{Evaluation sample for Task 2. Same symbol convention from Fig.\,\ref{fig:simple_task_visualization}. Note the increased number of false and missed measurements, coupled with more measurement and motion noise, making the tracking considerably harder.}
    \label{fig:complex_task_visualization}
\end{figure}
The resulting average GOSPA scores for Task 2 are shown in Table \ref{tab:gospa_scores_complex_task}, along with corresponding decompositions.
\begin{table}[b]
    \centering
   
    \caption{GOSPA and its decompositions for Task 2. \label{tab:gospa_scores_complex_task}}
    \begin{tabular}{@{}lllll@{}}
    \toprule
    \textbf{Method}  & \textbf{GOSPA} & \textbf{\begin{tabular}[c]{@{}l@{}}Localization\end{tabular}} & \textbf{\begin{tabular}[c]{@{}l@{}}Missed
    \end{tabular}} & \textbf{\begin{tabular}[c]{@{}l@{}}False\end{tabular}} \\ \midrule
    PMBM & 4.075 & 0.3025 & 3.225 & \textbf{0.163}
    \\
    $\delta$-GLMB & 4.450 & \textbf{0.280} & 3.515 & 0.323  \\
    MT3  & \textbf{3.662} & 0.3767 & \textbf{1.995} & 0.364
    \\
    \bottomrule
    \end{tabular}
\end{table}
In this setting, MT3 performs outperforms the alternatives, having not only considerably lower missed detection rates, but also the lowest overall GOSPA score. Traditional methods have more difficulty in keeping track of all possible data association hypotheses, and approximations must be made so that the posterior computation remains of feasible computational complexity \cite{mahler2014}. We hypothesize the Transformer model, in contrast, can efficiently learn to leverage complex time correlations in the input sequence to directly predict the object states, thus sidestepping the complicated posterior computation and resulting in better performance.

An example evaluation sample from this task is shown in Fig\, \ref{fig:complex_task_visualization}, together with the predictions for the best and second-best performing trackers. In it, we can see that the task is indeed harder than the previous one, but MT3 was still able to predict reasonable estimates for the position of most objects, whereas PMBM presented more missed detections, especially for newborn objects. For some detected objects (e.g., the one at approximately $(3, 7)$), although MT3 was able to detect it while PMBM failed to do so, MT3's state estimate seems suboptimal, judging from the trend in the measurements. This type of prediction may explain MT3's larger localization errors. From the results, we hypothesize that the Transformer architecture as used in MT3 is good at reasoning probable data association hypotheses, but future work could improve it for better regressing the state estimates.

\subsection{Ablation studies}
To shed light on the importance of the design choices for the MT3 algorithm, we conducted a series of ablation studies in Task 2. The following variants of the MT3 architecture were considered:
\begin{enumerate}
    \item \emph{No contrastive loss}: the term $\alpha$ in \eqref{eq:contrastive_loss} was set to zero.
    \item \emph{No iterative refinement}: $\mathbf \Delta_{1:k}^l$ in \eqref{eq:improved_measurements} was set to zeros for all $l\neq M$.
    \item \emph{No selection mechanism}: $\mathbf o_{1:k}$ in \eqref{eq:object_queries} is set to a learned lookup table (similar to \cite{DETR}), no dependence on $\mathbf z_{1:n}$ nor $\mathbf e_{1:n}$.
    \item \emph{No intermediate decoder losses}: $\mathcal L(\mathbf a_i^l, \mathbf g_j)$ in \eqref{eq:total_loss} is set to zero for all $l\neq M$.
\end{enumerate}
All other hyperparameters were kept constant, and the ablations were trained on exactly the same data, for the same number of gradient steps (600k). The average GOSPA scores for the original architecture and the ablations are shown in Table \ref{tab:ablation_studies}, computed using 1000 Monte Carlo simulations.
\begin{table}
    \centering
  
    \caption{Ablation studies for the MT3 architecture in Task 2.   \label{tab:ablation_studies}}
    \begin{tabular}{@{}ll@{}}
    \toprule
    \textbf{Ablation}  & \textbf{GOSPA} 
    \\ 
    \midrule
    MT3 (original) & 3.662
    \\
    MT3, no contrastive loss & 3.729
    \\
    MT3, no iterative refinement  & 4.094
    \\
    MT3, no selection mechanism & 4.587
    \\
    MT3, no intermediate decoder losses & 5.055
    \\
    \bottomrule
    \end{tabular}
    \vspace{-2mm}
\end{table}
The results show not only that all of the considered components are indeed beneficial in terms of final GOSPA score, but also their relative importance to final. Removing the intermediate decoder losses has the strongest effect, which agrees with other studies on the subject \cite{auxiliary_decoding_losses, DETR}, being a design decision of great importance when training deep Transformer decoders.

\section{Conclusion}
In this work, we performed a comparison of two state-of-the-art model-based Bayesian methods for multitarget tracking to a deep learning approach based on the Transformer architecture. The evaluation was performed in a simulated environment, assuming known multitarget models, and showed that the proposed architecture, MT3, performs competitively to the model-based Bayesian methods, while being able to outperform them in more complicated tasks. Our results provide evidence that deep-learning models are suitable for the MTT task also in the model-based regime, where model-based Bayesian trackers have long been regarded as state of the art. 

Interesting future research directions include developing training loss functions which are closer to the standard performance metric used in this setting (GOSPA) and investigating the performance of MT3 and related algorithms under model mismatch. Furthermore, we note that MT3 can be easily generalized to other contexts outside of the model-based regime without drastic changes to the overall architecture, and an investigation of its performance in new settings would shed more light into the power of Transformer-based architectures in MTT.

\section*{Acknowledgment}
This work was supported, in part, by a grant from the Chalmers AI Research Centre Consortium. Computational resources were provided by the Swedish National Infrastructure for Computing at C3SE, partially funded by the Swedish Research Council through grant agreement no. 2018-05973.

\bibliographystyle{IEEEtran}
\bibliography{refs}


\end{document}